\documentclass[10pt,twocolumn,letterpaper]{article}

\usepackage{iccv}              

\definecolor{iccvblue}{rgb}{0.21,0.49,0.74}
\usepackage[pagebackref,breaklinks,colorlinks,citecolor=iccvblue]{hyperref}
\usepackage{enumitem}
\usepackage{multirow}
\usepackage{colortbl}
\usepackage{makecell}
\usepackage{bm}
\newlist{mylist}{itemize}{1}
\setlist[mylist]{label=\textbullet}

\def\paperID{10541} 
\def\confName{ICCV}
\def\confYear{2025}

\title{HccePose(BF): Predicting Front \& Back Surfaces to Construct Ultra-Dense 2D-3D Correspondences for Pose Estimation}

\author{
	Yulin Wang \\
	Southeast University \\
	{\tt\small yulinwang@seu.edu.cn}
	\and
	Mengting Hu \\
	Southeast University \\
	{\tt\small 220240361@seu.edu.cn}
	\and
	Hongli Li \\
	Purdue University \\
	{\tt\small li5125@purdue.edu}
	\and
	Chen Luo\thanks{Corresponding author.} \\
	Southeast University \\
	{\tt\small chenluo@seu.edu.cn}
}

\begin{document}
\maketitle
\begin{abstract}
In pose estimation for seen objects, a prevalent pipeline involves using neural networks to predict dense 3D coordinates of the object surface on 2D images, which are then used to establish dense 2D-3D correspondences. However, current methods primarily focus on more efficient encoding techniques to improve the precision of predicted 3D coordinates on the object's front surface, overlooking the potential benefits of incorporating the back surface and interior of the object. To better utilize the full surface and interior of the object, this study predicts 3D coordinates of both the object's front and back surfaces and densely samples 3D coordinates between them. This process creates ultra-dense 2D-3D correspondences, effectively enhancing pose estimation accuracy based on the Perspective-n-Point (PnP) algorithm. Additionally, we propose Hierarchical Continuous Coordinate Encoding (HCCE) to provide a more accurate and efficient representation of front and back surface coordinates. Experimental results show that, compared to existing state-of-the-art (SOTA) methods on the BOP website, the proposed approach outperforms across seven classic BOP core datasets. Code is available at \url{https://github.com/WangYuLin-SEU/HCCEPose}.
\end{abstract}    
\section{Introduction}
\label{sec:Introduction}

\begin{figure}[t]
	\centering
	\includegraphics[width=0.85\linewidth]{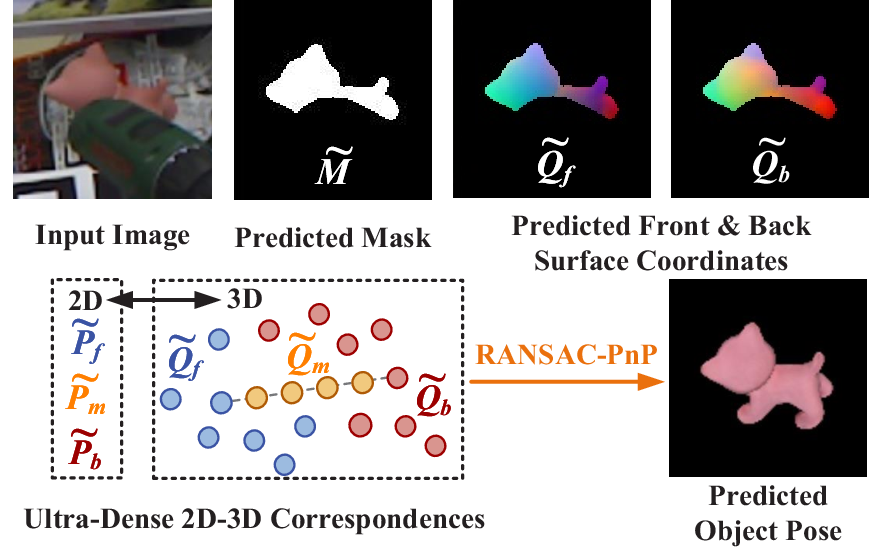}
	\caption{Pose estimation based on ultra-dense 2D-3D correspondences. For the input image, the network predicts the object mask and the coordinates of the object’s front and back surfaces. Based on the predicted coordinates, we perform dense sampling of 3D coordinates between the front and back surfaces to construct ultra-dense 2D-3D correspondences, and use the RANSAC-PnP solver \cite{epnp} to compute the object pose. Here, $\tilde{Q}_f$, $\tilde{Q}_b$, and $\tilde{Q}_m$ represent the sets of 3D coordinates for the object's front surface, back surface, and interior of the object, respectively. Their corresponding sets of 2D coordinates are denoted as $\tilde{P}_f$, $\tilde{P}_b$, and $\tilde{P}_m$.
	}
	\label{fig1:onecol}
\end{figure}

As a fundamental task in computer vision, pose estimation \cite{bib5,bib6} holds significant value across various applications, including automated product picking in warehouses \cite{bib81}, sheet metal assembly in industrial settings \cite{bib82}, and object tracking in virtual reality \cite{bib83}. In pose estimation for seen objects, some studies \cite{bib7,bib84,bib48,bib39,bib77,bib27,bib32,bib105} predefine 3D keypoints on the surface of a given object and use neural networks to predict the 2D projections of them on 2D images, establishing sparse 2D-3D correspondences. These correspondences are then used with the Perspective-n-Point (PnP) algorithm \cite{epnp} to compute the object pose.  Subsequently, some methods \cite{bib60,bib14,bib3,bib85,bib2,bib54,bib10,bib43,gdrnpp,repose} leverage neural networks to predict the 3D coordinates of the object's front surface from 2D images, constructing denser 2D-3D correspondences that significantly enhance the pose estimation accuracy achieved by PnP. However, existing studies typically focus only on the front surface of the object, overlooking the potential benefits of incorporating the back surface and interior of the object. Therefore, this paper proposes utilizing neural networks to simultaneously predict the coordinates of both the object’s front and back surfaces, along with densely sampling 3D coordinates between these surfaces, further enhancing the density of the 2D-3D correspondences. By leveraging this ultra-dense 2D-3D correspondence, the RANSAC-PnP solver can achieve more precise object poses (see Fig. \ref{fig1:onecol}).

The accuracy of object pose estimation using the PnP algorithm heavily depends on the precision of the predicted surface coordinates. To enhance the accuracy of both front and back surface coordinate predictions, this paper proposes an efficient Hierarchical Continuous Coordinate Encoding (HCCE). HCCE encodes $x$, $y$, and $z$ components of each surface coordinate separately and employs mirroring operations to transform these components into multi-level continuous codes. Neural networks are then used to predict these hierarchical continuous codes based on the input 2D image, enabling the calculation of the object’s front and back surface coordinates. Furthermore, during network training, our method calculates a separate histogram for each coordinate component, which records the proportion of mispredicted codes at different hierarchical levels. This reflects the learning difficulty of the network at various levels. Based on these error histograms, we adjust the weights of different hierarchical codes in the loss function, facilitating effective hierarchical learning. By utilizing weight adjustments based on multiple histograms rather than a single histogram \cite{bib10}, the stability of training and the accuracy of the predicted surface coordinates can be effectively enhanced.

Experimental results show that on the seven classic BOP core datasets \cite{bib82,bib60,bib5,tless,bib56,hb,bib29}, our method outperforms SOTA RGB-based approach \cite{bib10} by 2.4\% in BOP score. When trained on RGB but tested on RGB-D data, it achieves a 4.7\% improvement over the SOTA method \cite{gdrnpp}. Additionally, in the 2D segmentation task, our method surpasses the best existing approach \cite{bib10} by 3.7\%, further demonstrating its effectiveness.

Our contributions are summarized as follows:
\begin{mylist}
	\item To the best of our knowledge, this study is the first to demonstrate a neural network's ability to simultaneously predict both the front and back surfaces of an object for pose estimation.
	\item Our method constructs ultra-dense 2D-3D correspondences by densely and uniformly sampling 3D coordinates across the front and back surfaces, which effectively enhances pose accuracy.
	\item Our approach introduces a novel Hierarchical Continuous Coordinate Encoding (HCCE), which effectively enhances the accuracy of the predicted coordinates.
	\item We propose a hierarchical learning approach based on multiple histograms, which enables more efficient and stable training of networks based on HCCE.
\end{mylist}
\begin{figure*}[t]
	\centering
	\includegraphics[width=0.9\linewidth]{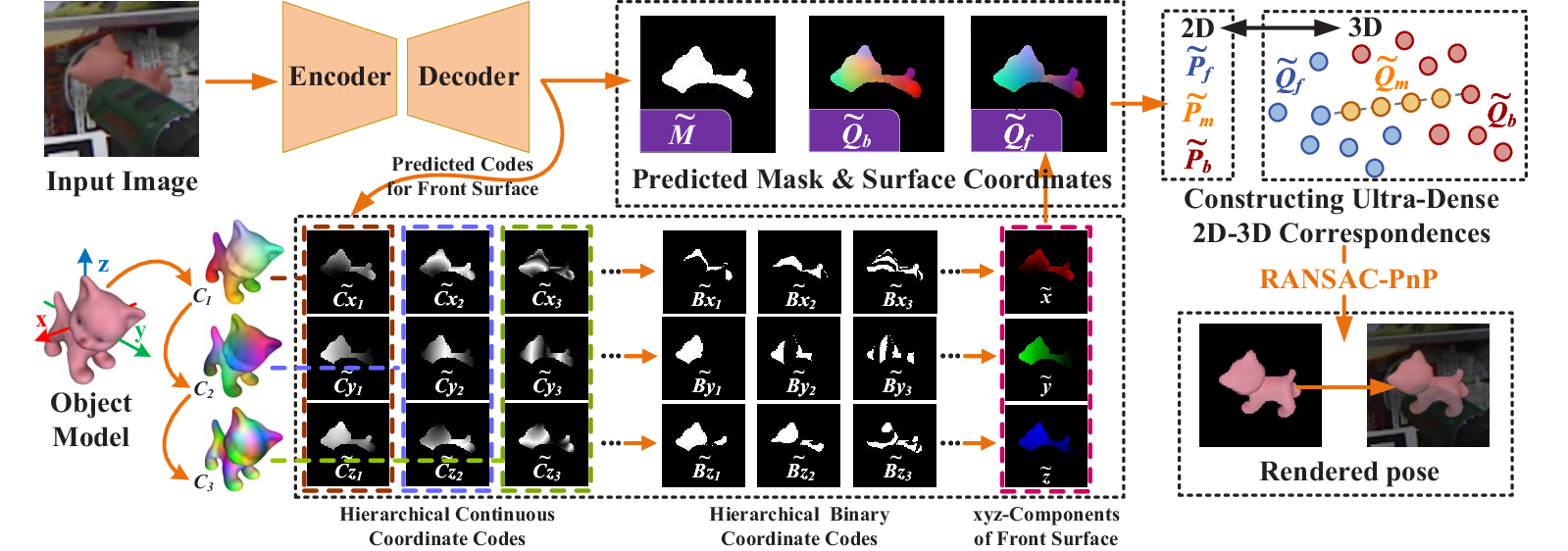}
	\caption{The inference pipeline. Our method begins by cropping a raw image based on the results of 2D object detection. The cropped image is then used as input to neural networks, which predict both the object mask and the coordinates of the front and back surfaces. To efficiently and accurately represent surface coordinates, we propose HCCE that encodes surface coordinates as multi-level codes, which are then learned by neural networks. During inference, the predicted multi-level continuous codes are first converted into binary codes, which are subsequently used to decode the surface coordinates (illustrated for the front surface coordinates in the figure). Using these predicted front and back surface coordinates, we densely sample 3D coordinates between the two surfaces to construct ultra-dense 2D-3D correspondences. Based on these ultra-dense correspondences, the method applies the RANSAC-PnP \cite{epnp} to compute the object’s pose.}
	\label{fig2:twocol}
\end{figure*}
\section{Related Work}
\label{sec:relatedwork}

To estimate the pose of seen objects, early methods \cite{bib56,bib29,bib69,bib68,bib72,bib73,bib74,bib75,bib76,bib101} directly regressed 3D translation and rotation using neural networks. However, such approaches are less robust under occlusion or background clutter. To overcome these limitations, later works \cite{bib7,bib84,bib48,bib39,bib77,bib27,bib32,bib105,bib3,bib2,bib54,bib10,bib43,gdrnpp,bib60,bib14,bib85} adopted a two-stage pipeline that first constructs 2D-3D correspondences and then computes the pose using a PnP algorithm \cite{epnp}, leading to more accurate pose estimation.

\subsection{PnP-based Object Pose Estimation}

In PnP-based object pose estimation, related work can be categorized into keypoint-based methods and surface coordinate-based methods. Keypoint-based methods \cite{bib7,bib84,bib48,bib39,bib77,bib27,bib32,bib105} typically predefine a set of 3D keypoints on the object model, such as the corner points of the object's 3D bounding box (BB8 \cite{bib7}) or keypoints selected using farthest point sampling \cite{fps1,fps2} (PvNet \cite{bib27}). These methods utilize neural networks to predict the 2D projections of these 3D keypoints on the input 2D image, thereby constructing sparse 2D-3D correspondences. To establish dense 2D-3D correspondences, surface coordinate-based methods \cite{bib3,bib2,bib43,bib54,bib10,repose,bib89,bib85,gdrnpp} predict the 3D coordinates of the object’s front surface  corresponding to each 2D pixel coordinate on the input 2D image. The poses computed via the PnP algorithm \cite{epnp} using these dense 2D-3D correspondences are generally more accurate than those derived from keypoint-based methods.

\subsection{Surface Coordinate Encoding}

To accurately predict the 3D coordinate of an object's surface, surface coordinate-based methods have proposed various encoding techniques, generally categorized as continuous encoding \cite{bib3,bib2,bib43} and discrete encoding \cite{bib54,bib10,repose}. In continuous encoding, methods such as CDPN \cite{bib3}, GDR-Net \cite{bib43}, and Pix2Pose \cite{bib2} normalize the surface coordinates and utilize neural networks to predict the normalized coordinates corresponding to 2D pixel coordinates on the input 2D image. In discrete encoding, methods like DPOD \cite{bib54} and ZebraPose \cite{bib10} encode different regions of the object surface into discrete codes and then use neural networks to predict the discrete codes corresponding to 2D pixel coordinates on the input image. Subsequently, these methods decode the surface regions using the predicted discrete codes, transforming them into surface coordinates. For instance,  when encoding object surface regions, DPOD \cite{bib54} employs a two-channel UV map to segment the surface into different regions, while ZebraPose \cite{bib10} uses clustering algorithms to encode the surface into hierarchical binary codes. This hierarchical encoding structure enables ZebraPose \cite{bib10} to achieve SOTA performance among RGB-based methods. 

The methods discussed above \cite{bib3,bib2,bib54,bib10,bib43,gdrnpp,repose} develop various encoding techniques to predict surface coordinates and have achieved accurate pose estimation. However, they share a common limitation: only the front surface of the object is considered for prediction. Although StereoPose \cite{StereoPose} also predicts both front and back surface coordinates, it is specifically designed for transparent objects and relies on stereo images. In contrast, our method simultaneously predicts both front and back surface coordinates, and performs dense sampling between them  to construct ultra-dense 2D-3D correspondences. Additionally, we introduce a effective encoding method, HCCE, which significantly outperforms traditional hierarchical binary encoding \cite{bib10} in surface coordinate prediction accuracy. Leveraging high-precision coordinates and ultra-dense 2D-3D correspondences, our approach achieves improved accuracy in pose estimation.

\section{Methodology}
\label{sec:Methodology}

Our method simultaneously predict the front and back surface coordinates through a neural network. Based on these predictions, dense sampling is performed to construct ultra-dense 2D-3D correspondences, leveraging potential benefits of the object's back and intermediate regions. These correspondences are then used by the RANSAC-PnP algorithm \cite{epnp} for pose estimation. To further enhance performance, we introduce HCCE to encode both surface coordinates, while a hierarchical learning approach improves training stability. The overall method is illustrated in Fig. \ref{fig2:twocol}.

Next, this section introduces the construction of ultra-dense 2D-3D correspondences and pose estimation (Section \ref{sec 3.1}), the Hierarchical Continuous Coordinate Encoding method (Section \ref{sec 3.2}), and the loss function for the neural network (Section \ref{sec 3.3}).

\subsection{Ultra-Dense 2D-3D Correspondences}
\label{sec 3.1}
To harness the potential benefits of incorporating the object's back surface and interior, we construct Ultra-Dense 2D-3D correspondences. The method first utilizes a neural network to simultaneously predict the 3D coordinates of both the object’s front and back surfaces, denoted as ${{\widetilde{Q}}_{f}}$ and ${{\widetilde{Q}}_{b}}$, respectively. Their corresponding 2D coordinates are represented as ${{\widetilde{P}}_{f}}$ and ${{\widetilde{P}}_{b}}$, where ${{\widetilde{P}}_{f}}={{\widetilde{P}}_{b}}$.

To further enrich the correspondences, we perform dense and uniform sampling between the front and back surfaces. Before sampling, a k-d tree is used to compute the average distance between the nearest point pairs in the predicted 3D coordinates, denoted as $\overline{d}$. Given the front and back 3D coordinates $\widetilde{{{q}_{1}}}$ and $\widetilde{{{q}_{2}}}$ corresponding to the same 2D coordinate, the number of interpolation points $n$ is determined by:
\begin{equation}
	n = \lfloor  \frac{\|\widetilde{q}_1 - \widetilde{q}_2\|_2}{\overline{d}} \rfloor
\end{equation}

Based on the sampling quantity $n$, uniform dense sampling between the front and back surfaces can be achieved using the following function:
\begin{equation}
	s\left( \widetilde{{{q}_{1}}},\widetilde{{{q}_{2}}},a \right)=a\times \widetilde{{{q}_{1}}}+\left( 1-a \right)\widetilde{{{q}_{2}}}
\end{equation}
where $a\in \left\{ \left. \frac{t}{n+1} \right|t=1,2,...,n \right\}$.

The resulting densely sampled 3D points are denoted as ${\widetilde{Q}_{m}}$ with corresponding 2D coordinates ${\widetilde{P}_{m}}$, where ${{\widetilde{P}}_{m}}={\widetilde{P}}_{f}={{\widetilde{P}}_{b}}$. Ultimately, the Ultra-Dense 2D-3D correspondences are constructed as:
\begin{equation}
	{{\widetilde{P}}_{u}}=\left\{ {{\widetilde{P}}_{b}},{{\widetilde{P}}_{f}},{{\widetilde{P}}_{m}} \right\}, {{\widetilde{Q}}_{u}}=\left\{ {{\widetilde{Q}}_{b}},{{\widetilde{Q}}_{f}},{{\widetilde{Q}}_{m}} \right\}
\end{equation}

After constructing ultra-dense 2D-3D correspondences, we observe that multiple 3D points share the same 2D projection (${{\widetilde{P}}_{m}}={\widetilde{P}}_{f}={{\widetilde{P}}_{b}}$). Directly applying RANSAC-PnP \cite{epnp} may therefore select multiple 3D points corresponding to the same 2D pixel in a single iteration, which can lead to unreliable pose estimates. To address this, we constrain each RANSAC-PnP iteration to sample only one 3D point per 2D pixel. The RANSAC process is repeated 150 times, and in each iteration, the reprojection error of the predicted front and back 3D points is computed using a 2-pixel threshold. The pose with the lowest error is finally selected as the result.


\subsection{Hierarchical Coordinate Encoding}
\label{sec 3.2}
\begin{figure}[t]
	\centering
	\includegraphics[width=0.9\linewidth]{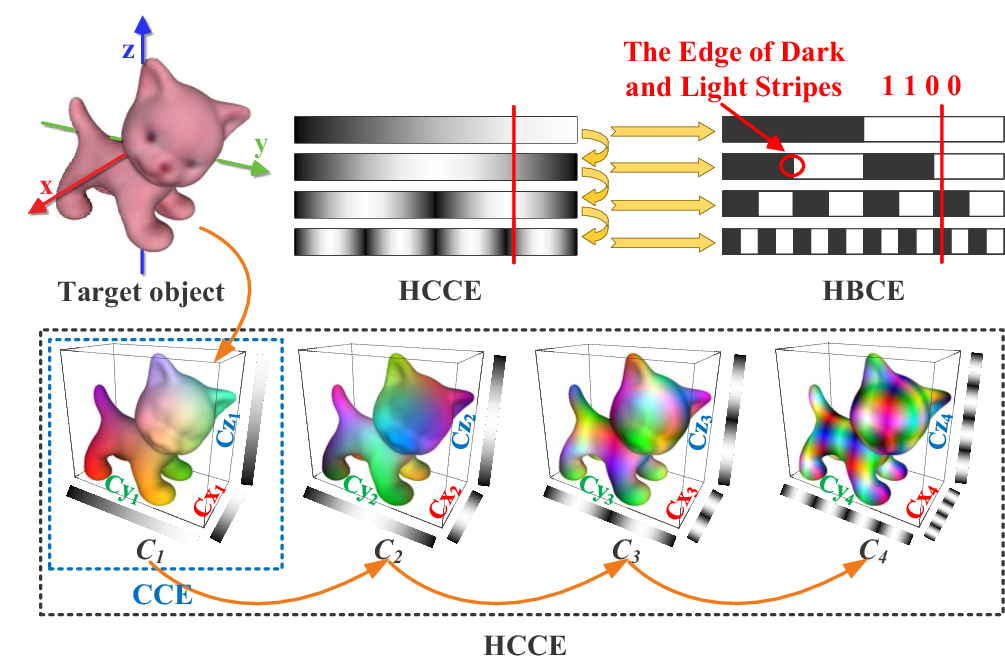}
	\caption{The hierarchical coordinate encoding method. In Hierarchical Binary Coordinate Encoding (HBCE), we observed that neural networks struggle to learn codes near the edges between dark and light stripes. To eliminate these stripes, we propose Hierarchical Continuous Coordinate Encoding (HCCE), which encodes coordinate components as multi-level continuous codes. Here, ${{C}_{1}}$ to ${{C}_{4}}$ represent the hierarchical continuous encoding of the object from the first to the fourth level, respectively. In addition, Continuous Coordinate Encoding (CCE) is an encoding method widely used by approaches such as CDPN \cite{bib3}, GDR-Net \cite{bib43}, and Pix2Pose \cite{bib2}.}
	\label{fig3:onecol}
\end{figure}

For encoding object surface coordinates, hierarchical surface encoding \cite{bib10} has demonstrated greater efficiency compared to non-hierarchical approaches \cite{bib3,bib2,bib43,bib54}. Existing approaches \cite{bib10} typically use hierarchical binary codes to represent different regions of an object’s surface. However, experiments in the related work have not explored whether it is more effective to encode surface regions or coordinate components. To investigate this, we introduce Hierarchical Binary Coordinate Encoding (HBCE), which we find improves pose estimation accuracy over hierarchical binary surface encoding. Nevertheless, during the training of HBCE-based methods, we observed that the neural network faced challenges in learning binary codes near the edges of dark and light stripes, as illustrated in Fig. \ref{fig3:onecol}. To address this, we propose Hierarchical Continuous Coordinate Encoding (HCCE), which utilizes multi-level continuous codes in place of binary codes to eliminate these stripes. Next, we will introduce HBCE and HCCE in detail.
\subsubsection{Hierarchical Binary Coordinate Encoding (HBCE)}

In HBCE, each component of a surface coordinate  is encoded as multi-level binary codes, where the coordinate components are normalized to the range of 0 to 1. The maximum of $k$ corresponds to the number of pixels occupied by the object in the input image. The traditional binary encoding function for floating-point numbers is employed to encode coordinate components. Based on the binary encoding function at level $i$, the components $x$, $y$, and $z$ of the surface coordinates can be encoded as binary code $B{{x}_{i,k}}$, $B{{y}_{i,k}}$, and $B{{z}_{i,k}}$, respectively, as illustrated in Fig. \ref{fig2:twocol}.

Using neural networks, multi-level binary codes are predicted from the input 2D image. These predicted binary codes enable the calculation of the surface coordinate components. For example, the decoding of the component $x$ can be expressed as:
\begin{equation}
	\label{eq2}
	{{x}_{k}}\approx \sum\limits_{i=1}^{8}{{{2}^{-k}}\times B{{x}_{i,k}}}
\end{equation}

Since excessively high encoding levels do not significantly enhance the precision of the represented coordinate components, this paper sets the upper limit of the encoding level to 8.
\subsubsection{Hierarchical Continuous Coordinate Encoding (HCCE)} 

In HCCE, hierarchical continuous codes are generated through mirroring operations. For surface coordinate prediction, the method first predicts the hierarchical continuous codes from the input image. The continuous codes are subsequently converted into binary codes, which are then used to calculate the surface coordinates, following a process consistent with HBCE. Next, we will describe the specific steps involved in HCCE and how the continuous codes are converted into binary codes. 

In encoding the surface coordinates, HCCE encodes each component of the coordinates into multi-level continuous codes. Given a coordinate component ${{x}_{k}}$, its continuous code at the first level is simply $C{{x}_{1,k}}={{f}_{1}}\left( {{x}_{k}} \right)={{x}_{k}}$. To ensure the continuity of the higher-level codes, the $\left( i-1 \right)$-th level code is mirrored to compute the $i$-th level code, where $i>1$, as follows:
\begin{equation}
	\label{eq3}
	C{{x}_{i,k}}={{f}_{i}}\left( {{x}_{k}} \right)=\begin{cases}
		{{f}_{i-1}}\left( 2{{x}_{k}} \right),{{x}_{k}}<0.5  \\
		{{f}_{i-1}}\left( 2-2{{x}_{k}} \right),{{x}_{k}}\ge 0.5 
	\end{cases} 
\end{equation} 
In this process, ${{f}_{i-1}}\left( \cdot  \right)$ and ${{f}_{i}}\left( \cdot  \right)$ are the continuous encoding functions for the $\left( i-1 \right)$-th and $i$-th levels, respectively. When ${{x}_{i}}<0.5$, the continuous code at the $i$-th level is obtained by copying the continuous code from the $\left( i-1 \right)$-th level; when ${{x}_{i}}\ge 0.5$, the continuous code at the $i$-th level is derived by mirroring the continuous code from the $\left( i-1 \right)$-th level.

In decoding the surface coordinates, the method first converts the continuous codes into binary codes and then calculates the surface coordinates according to Eq. (\ref{eq2}). In the conversion from continuous codes to binary codes, a binarization function $g\left( t \right)$ is defined such that $g\left( t \right)=0$ when $t<0.5$ and $g\left( t \right)=1$ when $t\ge 0.5$. Thus, the continuous code at the first level $C{{x}_{1,k}}$ can be converted to the binary code $B{{x}_{1,k}}=g\left( C{{x}_{1,k}} \right)$.

Subsequently, based on the binary code $B{{x}_{i-1,k}}$ from the $\left( i-1 \right)$-th level, the continuous code at the $i$-th level $C{{x}_{i,k}}$ can be converted to $B{{x}_{i,k}}$:
\begin{equation}
	\label{eq4}
	B{{x}_{i,k}}=\begin{cases}
		g\left( C{{x}_{i,k}} \right),B{{x}_{i-1,k}}=0  \\
		1-g\left( C{{x}_{i,k}} \right),B{{x}_{i-1,k}}=1  
	\end{cases} 
\end{equation}
Here, $B{{x}_{i-1,k}}=1$ indicates that $C{{x}_{i,k}}$ was generated through mirroring, which means that the binarized $g\left( C{{x}_{i,k}} \right)$ does not correspond to the hierarchical binary code $B{{x}_{i,k}}$. Therefore, mirroring is reversed in the decoding process, resulting in $B{{x}_{i,k}}=1-g\left( C{{x}_{i,k}} \right)$.
\subsection{The Loss Function of Neural Network} 
\label{sec 3.3}

In neural network training, the proposed method utilizes a neural network to learn the object mask and the hierarchical continuous codes of front and back surface coordinates. With the number of encoding levels for each coordinate component set to 8, the neural network outputs 49 channels, comprising 3×8 channels for the front surface coordinates, 3×8 channels for the back surface coordinates, and 1 channel for the object mask. Consequently, the loss function of networks can be divided into mask loss ${{L}_{M}}$ and hierarchical loss (front loss $L_{xyz}^{Front}$ and back loss $L_{xyz}^{Back}$). The total loss of the neural network can be expressed as: $L={{L}_{M}}+\gamma \times (L_{xyz}^{Front}+L_{xyz}^{Back})$, where $\gamma $ is the weighting coefficient for the hierarchical loss in the total loss function.
\subsubsection{Mask Loss}  
\label{sec3.3.1}
During training, the proposed method utilizes the L1 metric to compute the loss ${{L}_{M}}$ between the ground-truth object mask $M$ and the predicted object mask $\widetilde{M}$, ${{L}_{M}}=\sum\limits_{{}}^{m}{{{\left\| {{M}_{i}}-\widetilde{{{M}_{i}}} \right\|}_{1}}}$. $\widetilde{{{M}_{i}}}$ denotes the output on a pixel and $m$ describes the number of outputs. Pixels with $\widetilde{M} > 0$ correspond to the predicted object mask, while those with $\widetilde{M} \leq 0$ are considered background. The total number of pixels classified as the object mask is denoted as $n$.
\subsubsection{Hierarchical Loss}  
\label{sec3.3.2}
During training, the proposed method uses histograms to quantify the proportion of erroneous predictions at different hierarchical levels. This approach effectively measures the difficulty the neural network encounters in learning across various levels. Based on these error histograms, the method dynamically adjusts the weights of the continuous codes in the loss function at each level. This adjustment facilitates hierarchical learning and enhances the stability of neural network training. 

For different coordinate components, the method computes the corresponding histograms separately. For instance, consider the calculation of the histogram ${{h}_{f,x}}$ for the $x$ component of the front surface coordinates. The process begins by obtaining the predicted hierarchical continuous code ${{\widetilde{Cx}}_{_{f,i}}}$ at the $i$-th level, which is then converted into the corresponding hierarchical binary code ${{\widetilde{Bx}}_{_{f,i}}}$. Next, the proportion of erroneous binary codes at the $i$-th level, denoted as ${{r}_{f,}}_{x,i}$, is calculated. Consequently, the intensity ${{h}_{f,x,i}}$ of the $x$ component in the histogram ${{h}_{f,x}}$ at the $i$-th level can be computed as follows:
\begin{equation}
	\label{eq5}
	{{h}_{f,x,i}}=\exp \big( \sigma \cdot \min \left\{ {{r}_{f,x,i}},0.5-{{r}_{f,x,i}} \right\} \big)
\end{equation}
To reduce the weight of codes at levels with a high proportion of errors in the loss function, ${{r}_{f,x,s}}$ is truncated at 0.5. An exponential term $\exp \left( \cdot  \right)$ is used to smooth the abrupt changes in intensity caused by the truncation. Additionally, $\sigma $ is a constant employed to adjust the intensity range in the histogram. By normalizing ${{h}_{f,x,i}}$ in the histogram ${{h}_{f,x}}$, the weight ${{w}_{f,x,i}}$ for the $i$-th level in the loss function can be obtained. 

Based on the weights ${{w}_{f,x,i}}$ for each level, the loss function $L_{x}^{Front}$ can be expressed as follows:
\begin{equation}
	\label{eq6}
	L_{x}^{Front}=\sum\limits_{i=1}^{8}{\left( {{w}_{f,x,i}}\cdot \sum\limits_{j=1}^{n}{{{\left\| C{{x}_{f,i,j}}-{{\widetilde{Cx}}_{_{f,i,j}}} \right\|}_{1}}} \right)}
\end{equation} 
Here, ${{\widetilde{Cx}}_{f,i,j}}$ denotes the $j$-th code at the $i$-th level, where the upper limit for $j$ is the area $n$ of the predicted mask $\widetilde{M}$. 

Similarly, the histograms for the $y$ and $z$ components of the object’s front surface, along with the histogram-based loss functions $L_{y}^{Front}$ and $L_{z}^{Front}$, can also be calculated. Thus, the hierarchical loss for the object’s front surface can be expressed as $L_{xyz}^{Front}=L_{x}^{Front}+L_{y}^{Front}+L_{z}^{Front}$. Likewise, the hierarchical loss $L_{xyz}^{Back}$ for the object’s back surface can be computed in the same manner.


\section{Experiments}
\label{sec:Experiments}
In this section, we first introduce the experimental setup, including implementation details, datasets, and evaluation metrics. Next, we conduct ablation studies on the encoding method, loss function weight adjustment, and 2D-3D correspondences. Finally, we compare our experimental results with SOTA methods \cite{bib10,gdrnpp} on the BOP website across seven classic BOP core datasets \cite{bib82,bib60,bib5,tless,bib56,hb,bib29}.

\subsection{Experimental Setup}

\subsubsection{Implementation Details}

\textbf{Network Setup.} The network architecture in this paper is adapted from ZebraPose \cite{bib10}, with a ResNet34 \cite{ResNet34} version for ablation experiments and an EfficientNet-B4 \cite{Efficientnet} version for comparison experiments. Both network versions take RGB images as input and output the object mask along with the 3D surface coordinates. The input and output resolutions are 256×256 and 128×128, respectively.

\textbf{Preprocessing of Training Data.} The DZI strategy, proposed by CDPN \cite{bib3}, is employed to simulate randomly detected 2D bounding boxes. Additionally, the random enhancement strategy, introduced by GDR-Net \cite{bib43}, is incorporated to simulate real-world RGB image noise.

\textbf{Training Setup.} Separate neural networks are trained for each object, with each training process taking approximately 24 hours on a single NVIDIA RTX 4090 GPU. For the ablation experiments, the ResNet34-based version is used, with a maximum of 100k training epochs, a batch size of 24, and a constant learning rate of 0.0002. For the comparative experiments, the EfficientNet-B4 version is used, with a maximum of 600k training epochs and a batch size of 12. Pose estimation for a single object within a bounding box takes approximately 30 ms during inference. 

\textbf{Label Preparation.} Following the BOP toolkit’s practice of rendering front surface depth maps using VisPy to generate coordinate labels, we adopt a similar approach with PyOpenGL. By setting the depth test to \texttt{GL\_LESS} and \texttt{GL\_GREATER}, we respectively obtain front and back surface depth maps, which are then used to generate the corresponding labels.
\subsubsection{Datasets}

The performance of the method was tested on seven classic BOP core datasets: LM-O \cite{bib60}, YCB-V \cite{bib29}, IC-BIN \cite{icbin}, TUD-L \cite{bib6}, HB \cite{hb}, T-LESS \cite{tless}, and ITODD \cite{bib82}. During training, we used the PBR \cite{blenderproc,bib31} training data provided by BOP \cite{bop2023,bop2022} for all seven datasets. For the YCB-V, T-LESS, and TUD-L datasets, real data was also used in training. During testing, we utilized the detection sequences and 2D detection results provided by the BOP 2023 challenge \cite{bop2023,bop2022}.

\subsubsection{Evaluation Metrics}
\label{sec4.1.3}
\textbf{ADD(-S) Accuracy} \cite{lmo}\textbf{.} In the ablation experiments, we use the widely adopted ADD(-S) metric to evaluate the accuracy of object pose predictions. It includes ADD for non-symmetric objects and ADD-S for rotationally symmetric objects. For objects without rotational symmetry, ADD computes the average distance between the transformed vertices of the object based on the predicted pose and the ground truth pose. For objects with rotational symmetry, ADD-S calculates the average distance between the nearest vertices, rather than the same vertex. A predicted pose is considered correct if the ADD(-S) error is less than 10\% of the object's diameter. ADD(-S) accuracy, also referred to as average recall (AR), represents the proportion of correctly predicted poses across all test sequences. Additionally, we also report the Area Under the Curve (AUC) for ADD(-S) and ADD-S \cite{bib29}, with a maximum threshold of 10 cm.

\textbf{Coordinate Accuracy.} In addition to reporting pose accuracy, the ablation experiments also assess the accuracy of predicted surface coordinates. A prediction is considered correct if the Euclidean distance between the predicted and ground truth coordinates is below a given threshold. Coordinate accuracy is defined as the ratio of correctly predicted coordinates to the total number of predicted coordinates. 
 
\textbf{BOP Score.} In the comparison experiments, the BOP score of the proposed method is presented on seven classic BOP core datasets \cite{bib82,bib60,bib5,tless,bib56,hb,bib29}. The BOP score is the average of three evaluations: VSD (Vertex Symmetric Distance), MSSD (Mean Symmetric Surface Distance), and MSPD (Mean Symmetric Projection Distance) \cite{bib5,bib6}. These evaluations have complex definitions, and the specific details can be found on the BOP website.

\subsection{Ablation Experiments}

Ablation experiments on encoding methods and loss weight adjustments were conducted on the IC-BIN dataset \cite{icbin}. The results, presented in Fig. \ref{fig5:onecol} and Tab. \ref{tab1}, include the prediction accuracy of object surface coordinates and object pose accuracy, with a primary focus on ADD(-S) accuracy. Additionally, ablation studies on 2D-3D correspondences were performed on three datasets: LM-O \cite{bib60}, TUD-L \cite{bib6}, and IC-BIN \cite{icbin}, with the average precision (AP) of BOP scores reported in Tab. \ref{tab2}. To ensure fair comparisons, OpenCV’s RANSAC-PnP \cite{epnp} is consistently used across all methods, regardless of the original PnP algorithms employed by other approaches.

\subsubsection{Encoding Methods}

\begin{figure}[b]
	\centering
	\includegraphics[width=0.9\linewidth]{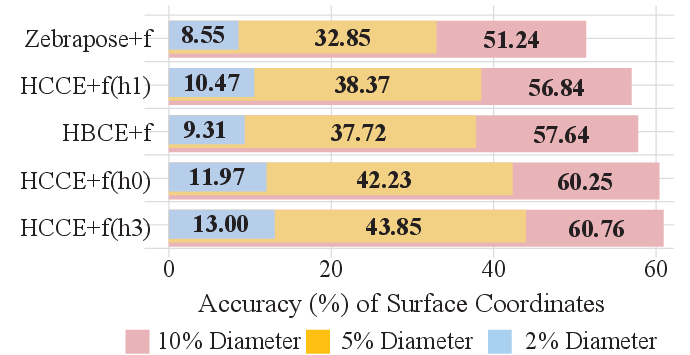}
	\caption{Accuracy of surface coordinates. Percentage of correctly predicted coordinates under thresholds of 2\%, 5\%, and 10\% of the object diameter.}
	\label{fig5:onecol}
\end{figure}

The current SOTA RGB-based pose estimation method, ZebraPose \cite{bib10}, employs hierarchical binary surface encoding, which represents different regions of the object surface as multi-level binary codes. However, we find that directly encoding coordinate components is more efficient and accurate than encoding surface regions. Experimental results show that our proposed hierarchical binary surface encoding (HBCE + f) improves surface coordinate accuracy (see Fig. \ref{fig5:onecol}) and enhances pose estimation accuracy by 0.97\% (see Tab. \ref{tab1}). However, it was observed that the network struggled to learn the binary codes near the edges of the dark and light stripes, as shown in Fig. \ref{fig3:onecol}. To eliminate these stripes, this paper introduces hierarchical continuous coordinate encoding (HCCE), which encodes coordinate components as multi-level continuous codes. Compared to HBCE, HCCE (HCCE + f(h3)) further improves the prediction accuracy of object surface coordinates and increases pose estimation accuracy by 5.13\%. This indicates that the neural network finds it easier to learn continuous codes than binary codes, which effectively enhances the accuracy of pose estimation.

\begin{table}[t]
	\centering
	\footnotesize
	\caption{The ablation experiments on the IC-BIN dataset \cite{icbin}.
		h0 signifies the absence of histogram-based weight adjustment; h1 denotes weight adjustment using a single histogram; h3 represents weight adjustment using multiple histograms; f indicates the prediction of only the object’s front surface. Cell colors indicate column-wise values from low (dark blue) to high (dark red).}
	\begin{tabular}{l|c|cccc}
		\toprule
		\multicolumn{2}{c}{Method}     & \makecell{AR of \\ ADD-(S)} & \makecell{AR of \\ ADD-S} & \makecell{AUC of \\ ADD(-S)} & \makecell{AUC of \\ ADD-S} \\
		\midrule
		\makecell{Zebra- \\ Pose}      & f         & \cellcolor[HTML]{D0E8F0} 55.85 & \cellcolor[HTML]{ADD8E6} 61.82 & \cellcolor[HTML]{ADD8E6} 72.91 & \cellcolor[HTML]{ADD8E6} 76.40 \\
		\midrule
		CCE  & f      & \cellcolor[HTML]{ADD8E6} 55.42 & \cellcolor[HTML]{D0E8F0} 62.15 & \cellcolor[HTML]{E8F4F8} 73.58 & \cellcolor[HTML]{E8F4F8} 78.14 \\
		\midrule
		HBCE & f      & \cellcolor[HTML]{E8F4F8} 56.82 & \cellcolor[HTML]{E8F4F8} 63.50 & \cellcolor[HTML]{D0E8F0} 73.15 & \cellcolor[HTML]{D0E8F0} 76.56 \\
		\midrule
		\multirow{3}{*}{HCCE} & f(h0)  & \cellcolor[HTML]{F0D0D0} 61.35 & \cellcolor[HTML]{F0D0D0} 67.00 & \cellcolor[HTML]{F0D0D0} 76.35 & \cellcolor[HTML]{F0D0D0} 79.82 \\
		& f(h1) & \cellcolor[HTML]{F8E8E8} 60.44   & \cellcolor[HTML]{F8E8E8} 65.23 & \cellcolor[HTML]{F8E8E8} 74.92 & \cellcolor[HTML]{F8E8E8} 78.68 \\
		& f(h3) & \cellcolor[HTML]{EAA4A4} 61.95  & \cellcolor[HTML]{EAA4A4} 68.30 & \cellcolor[HTML]{EAA4A4} 77.67 & \cellcolor[HTML]{EAA4A4} 81.11 \\
		\bottomrule
	\end{tabular}
	\label{tab1}
\end{table}
\subsubsection{Weight Adjustment}

\begin{figure}[t]
	\centering
	\includegraphics[width=0.9\linewidth]{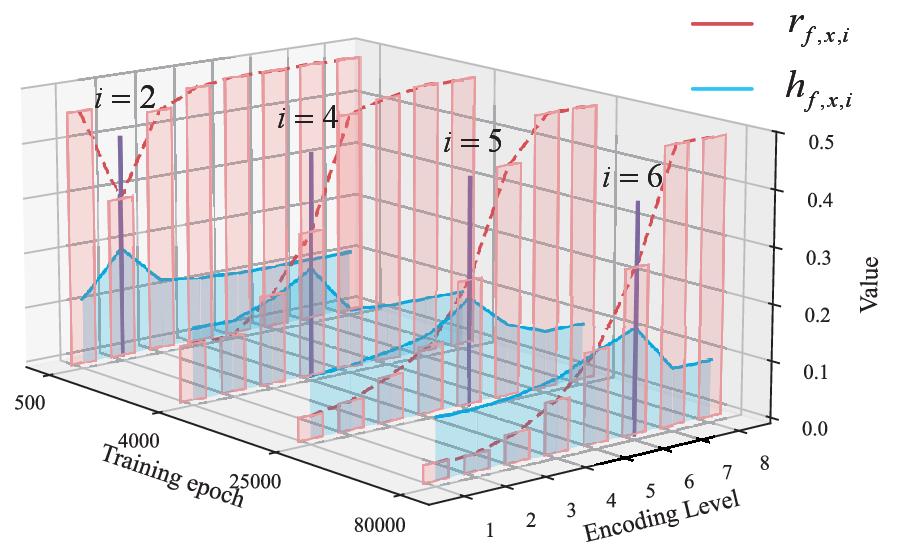}
	\caption{Weight adjustments across different training epochs.}
	\label{fig5.5:onecol}
\end{figure}

In hierarchical learning, higher-level codes are more complex; for example, ${{C}_{4}}$ in Fig. \ref{fig3:onecol} is much more complex than ${{C}_{1}}$. The more complex the code, the more difficult it is for neural networks to learn. To measure the learning difficulty at different levels, ZebraPose \cite{bib10} uses a single histogram to track the proportion of incorrect codes at each level. Subsequently, ZebraPose adjusts the weights of different levels in the loss function to facilitate hierarchical learning. However, hierarchical learning based on a single histogram is not suitable for HCCE. Compared to not adjusting the weights (HCCE + f(h0)), using dynamic weights based on a single histogram (HCCE + f(h1)) reduces the prediction accuracy of object surface coordinates, resulting in a 0.91\% decrease in ADD(-S) accuracy. Therefore, our method generates a separate histogram for each coordinate component and proposes dynamic weights based on multiple histograms. In Fig. \ref{fig5.5:onecol}, we illustrate the histogram intensities and weight coefficients at different encoding levels for the $x$-component of the 3D coordinates on the front surface. As training progresses, the peak of the weight coefficients shifts from lower to higher encoding levels, enabling hierarchical learning from low- to high-level encodings. Weight adjustment based on multiple histograms (HCCE + f(h3)) achieves more accurate coordinate predictions, improving the ADD(-S) accuracy by 0.6\% compared to not adjusting the weights. This demonstrates that weight adjustment based on multiple histograms can enhance the stability of network training and improve the accuracy of pose estimation.

\begin{table}[b]
	\centering
	\footnotesize
	\caption{Effect of front and back surface information on the Average Precision (AP) of BOP scores. bf denotes the simultaneous prediction of both the object's front and back surfaces, while bfu refers to joint prediction of the object's front and back surfaces, along with uniform sampling between them. Cell colors indicate row-wise values from low (dark blue) to high (dark red).}
	\begin{tabular}{c|c|cccc}
		\toprule
		\multicolumn{1}{l|}{Dataset}                  & Object                           & f    & b    & bf   & bfu  \\ 
		\midrule
		\multicolumn{1}{c|}{\multirow{8}{*}{LM-O \cite{bib60}}}   & \multicolumn{1}{c|}{ape}         & \cellcolor[HTML]{D0E8F0} 84.9 & \cellcolor[HTML]{F8E8E8} 85.7 & \cellcolor[HTML]{F8E8E8} 85.7 & \cellcolor[HTML]{F0D0D0} 85.8 \\
		\multicolumn{1}{c|}{}                        & \multicolumn{1}{c|}{can}         & \cellcolor[HTML]{D0E8F0} 94.8 & \cellcolor[HTML]{D0E8F0} 94.8 & \cellcolor[HTML]{F8E8E8} 95.2 & \cellcolor[HTML]{F0D0D0} 95.5 \\
		\multicolumn{1}{c|}{}                        & \multicolumn{1}{c|}{cat}         & \cellcolor[HTML]{E8F4F8} 87.2 & \cellcolor[HTML]{D0E8F0} 85.9 & \cellcolor[HTML]{F8E8E8} 87.6 & \cellcolor[HTML]{F0D0D0} 87.7 \\
		\multicolumn{1}{c|}{}                        & \multicolumn{1}{c|}{driller}     & \cellcolor[HTML]{D0E8F0} 93.1 & \cellcolor[HTML]{D0E8F0} 93.1 & \cellcolor[HTML]{F0D0D0} 94.5 & \cellcolor[HTML]{F0D0D0} 94.5 \\
		\multicolumn{1}{c|}{}                        & \multicolumn{1}{c|}{duck}        & \cellcolor[HTML]{D0E8F0} 78.7 & \cellcolor[HTML]{F8E8E8} 82.1 & \cellcolor[HTML]{E8F4F8} 80.5 & \cellcolor[HTML]{F0D0D0} 82.4 \\
		\multicolumn{1}{c|}{}                        & \multicolumn{1}{c|}{eggbox}      & \cellcolor[HTML]{E8F4F8} 56.6 & \cellcolor[HTML]{F0D0D0} 57.1 & \cellcolor[HTML]{D0E8F0} 56.0 & \cellcolor[HTML]{F0D0D0} 57.1 \\
		\multicolumn{1}{c|}{}                        & \multicolumn{1}{c|}{glue}        & \cellcolor[HTML]{E8F4F8} 84.4 & \cellcolor[HTML]{D0E8F0} 83.9 & \cellcolor[HTML]{F8E8E8} 85.8 & \cellcolor[HTML]{F0D0D0} 86.1 \\
		\multicolumn{1}{c|}{}                        & \multicolumn{1}{c|}{holepuncher} & \cellcolor[HTML]{F8E8E8} 82.2 & \cellcolor[HTML]{D0E8F0} 75.8 & \cellcolor[HTML]{E8F4F8} 80.1 & \cellcolor[HTML]{F0D0D0} 82.3 \\
		\midrule
		\multicolumn{1}{l|}{\multirow{3}{*}{TUD-L \cite{bib6}}}  & \multicolumn{1}{c|}{obj01}       & \cellcolor[HTML]{D0E8F0} 95.0 & \cellcolor[HTML]{E8F4F8} 95.4 & \cellcolor[HTML]{F8E8E8} 95.7 & \cellcolor[HTML]{F0D0D0} 95.8 \\
		\multicolumn{1}{l|}{}                        & \multicolumn{1}{c|}{obj02}       & \cellcolor[HTML]{D0E8F0} 85.4 & \cellcolor[HTML]{F8E8E8} 86.3 & \cellcolor[HTML]{E8F4F8} 86.1 & \cellcolor[HTML]{F0D0D0} 86.8 \\
		\multicolumn{1}{l|}{}                        & \multicolumn{1}{c|}{obj03}       & \cellcolor[HTML]{F8E8E8} 93.2 & \cellcolor[HTML]{D0E8F0} 92.2 & \cellcolor[HTML]{E8F4F8} 92.6 & \cellcolor[HTML]{F0D0D0} 93.3 \\
		\midrule
		\multicolumn{1}{l|}{\multirow{2}{*}{IC-BIN \cite{icbin}}} & \multicolumn{1}{c|}{obj01}       & \cellcolor[HTML]{E8F4F8} 62.9 & \cellcolor[HTML]{D0E8F0} 58.5 & \cellcolor[HTML]{F8E8E8} 63.6 & \cellcolor[HTML]{F0D0D0} 63.7 \\
		\multicolumn{1}{l|}{}                        & \multicolumn{1}{c|}{obj02}       & \cellcolor[HTML]{E8F4F8} 70.1 & \cellcolor[HTML]{D0E8F0} 65.9 & \cellcolor[HTML]{F8E8E8} 70.7 & \cellcolor[HTML]{F0D0D0} 71.5 \\
		\midrule
		\multicolumn{2}{c|}{Mean}                                                       & \cellcolor[HTML]{E8F4F8} 82.2 & \cellcolor[HTML]{D0E8F0} 81.3 & \cellcolor[HTML]{F8E8E8} 82.6 & \cellcolor[HTML]{F0D0D0} 83.3 \\
		\bottomrule
	\end{tabular}
	\label{tab2}
\end{table}

\begin{table*}[t]
	\centering
	\footnotesize
	\caption{Comparison with SOTA methods on seven classic BOP core datasets in 6D localization and 2D segmentation tasks. Scores are expressed as percentages.}
	\begin{tabular}{@{}l|ccc|ccc|ccc@{}}
		\toprule
		Task  & \multicolumn{3}{c|}{6D Localization} & \multicolumn{3}{c|}{2D Segmentation} & \multicolumn{3}{c}{6D Localization} \\
		\midrule
		Test Data & \multicolumn{3}{c|}{RGB} & \multicolumn{3}{c|}{RGB} & \multicolumn{3}{c}{RGB-D} \\
		\midrule
		Method & \multicolumn{1}{l}{GPose \cite{gpose}} & \multicolumn{1}{l}{ZebraPose \cite{bib10}} & \multicolumn{1}{l|}{Ours} & \multicolumn{1}{l}{DLZDet \cite{DLZDet}} & \multicolumn{1}{l}{ZebraPose \cite{bib10}} & \multicolumn{1}{l|}{Ours} & \multicolumn{1}{l}{PFA \cite{PFA}} & \multicolumn{1}{l}{GDRNPP \cite{gdrnpp}} & \multicolumn{1}{l}{Ours} \\
		\midrule
		${AR(AP)_{LM-O}}$ & 69.9  & 72.9  & \textbf{75.5}  & 46.0    & 51.6  & \textbf{52.7}  & 79.2  & 79.2  & \textbf{80.5} \\
		${AR(AP)_{T-LESS}}$ & 79.9  & 82.1  & \textbf{85.6}  & 58.4  & 72.1  & \textbf{76.8}  & 84.9  & 87.2  & \textbf{87.9} \\
		${AR(AP)_{TUD-L}}$ & 83.1  & 85.0    & \textbf{86.9}  & 73.2  & 71.8  & \textbf{76.7}  & \textbf{96.3}  & 93.6  & 94.4 \\
		${AR(AP)_{IC-BIN}}$ & 62.6  & 59.2  & \textbf{63.5}  & 31.6  & 49.3  & \textbf{51.2}  & 70.6  & 70.2  & \textbf{72.4} \\
		${AR(AP)_{ITODD}}$ & 46.0    & 50.4  & \textbf{54.2}  & 23.9  & 46.2  & \textbf{51.3}  & 52.6  & 58.8  & \textbf{73.4} \\
		${AR(AP)_{HB}}$ & 87.6  & \textbf{92.2}  & 91.9  & 60.0    & 68.9  & \textbf{76.1}  & 86.7  & 90.9  & \textbf{93.1} \\
		${AR(AP)_{YCB-V}}$ & 80.9  & 82.8  & \textbf{83.9}  & 66.9  & 73.1  & \textbf{74.5}  & 89.9  & 83.4  & \textbf{91.1} \\
		\midrule
		${AR(AP)_{Core}}$ & 72.9  & 74.9  & \textbf{77.3}  & 51.4  & 61.9  & \textbf{65.6}  & 80.0    & 80.5  & \textbf{84.7} \\
		\bottomrule
	\end{tabular}
	\label{tab3}
\end{table*}

\subsubsection{2D-3D correspondences}
\label{sec.4.2.3}
In this section, we examine the average precision (AP) of BOP scores for all objects in three datasets (LM-O \cite{bib60}, TUD-L \cite{bib6}, and IC-BIN \cite{icbin}) to assess how different surface information impacts pose estimation accuracy.

Tab. \ref{tab2} presents the impact on the AP of BOP scores across three datasets when 2D-3D correspondences are constructed based on the front surface, the back surface, and the intermediate 3D coordinates sampled between them. In the LM-O dataset, the contributions of front and back surface information to the improvement in BOP scores are roughly equivalent; in the TUD-L dataset, front surface information is less effective than back surface information; whereas in the IC-BIN dataset, front surface information proves more advantageous. These experimental results indicate that pose estimation based on front surface information and that based on back surface information each have their own strengths and weaknesses. Therefore, when using only front surface information yields unsatisfactory performance, incorporating back surface information can effectively enhance the accuracy of pose prediction.

Furthermore, the experimental results (see the “Mean” row in Tab. \ref{tab2}) show that simultaneously leveraging both front and back surface information increases the BOP scores by 0.4\% and 1.3\%, respectively. Among the 13 objects evaluated, 8 objects achieved BOP scores that were equal to or higher when using both surfaces compared to using only front or only back information. This suggests that for most objects, a hybrid approach that combines 2D-3D correspondences from both the front and back surfaces is more effective in extracting predictive information, thereby improving pose estimation performance.

To further capitalize on the predicted front and back surface coordinates, we uniformly sample new 3D coordinates between the two surfaces to construct ultra-dense 2D-3D correspondences. Compared to using only front or back surface information, the adoption of ultra-dense correspondences improves the BOP scores by 1.1\% and 2.0\%, respectively, further demonstrating that this approach more effectively integrates information from both surfaces, thereby enhancing the accuracy of pose estimation.

\subsection{Comparison with the SOTA Methods}

The comparison experiments were conducted on the seven classic BOP core datasets \cite{bib82,bib60,bib5,tless,bib56,hb,bib29}, and the BOP scores of our method can be viewed on the BOP website. To ensure a fair comparison, only methods based on the BOP 2023 official detection results were considered. The scores for the methods listed in Tab. \ref{tab3}, including ZebraPose \cite{bib10}, GPose \cite{gpose}, MegaPose \cite{megapose}, and GDRNPP \cite{bib43,gdrnpp}, are also obtained from the BOP website. Notably, if a method achieves higher scores with alternative detection results, those scores are not reported here.

When tested on RGB data, our method outperforms the best existing approach \cite{bib10} by 2.4\% in the BOP score, as shown in Tab. \ref{tab3}. Among the methods trained on RGB data but tested with RGB-D data, the current top-performing approach refines poses using Iterative Closest Point (ICP \cite{icp}). In contrast, our method leverages the more efficient FoundationPose \cite{bib86}, leading to a 4.7\% improvement in the BOP score (see Tab. \ref{tab3}). Furthermore, in the 2D segmentation task, our method surpasses existing approaches \cite{bib10} by 3.7\%, further demonstrating its effectiveness.

\section{Conclusion}


In this study, we propose an ultra-dense 2D-3D correspondence to enable more accurate pose estimation via RANSAC-PnP. It is constructed by simultaneously predicting the coordinates of both the front and back object surfaces, followed by dense sampling of intermediate 3D points. We further introduce an efficient Hierarchical Continuous Coordinate Encoding (HCCE) and a weight adjustment approach based on multiple histograms, which together facilitate accurate predictions of both surfaces. Compared to SOTA methods \cite{bib10, gdrnpp}, our method achieves competitive scores and rankings on the BOP Challenge. However, due to varying learning difficulties across objects, the trained network is inherently object-specific, making it infeasible to train a single unified model for multiple objects.

\section*{Acknowledgments}
This work was supported by the National Natural Science Foundation of China under grant  No. 52375487 and the special fund of Jiangsu Province for Key Research and Development No. BE2023041. Additionally, the third author carried out this work during her undergraduate and master’s studies at Southeast University under the supervision of Dr. Luo, prior to commencing her doctoral studies at Purdue University.

{
	\small

	\bibliographystyle{unsrt}
	\bibliography{main}
}


\end{document}